\documentclass[12pt,a4paper]{cibb}
\usepackage{fancyhdr}
\usepackage{subfigure,graphicx}
\usepackage{amsmath,amsfonts,latexsym,amssymb,euscript,xr}
\usepackage{booktabs}
\usepackage[nodayofweek]{datetime}
\usepackage{hyperref}
\usepackage[table]{xcolor}
\usepackage{color,colortbl,tabularx}
\usepackage[english]{babel}
\usepackage[protrusion=true,expansion=true]{microtype}
\usepackage{amsmath,amsfonts,amsthm}
\usepackage{pifont}
\usepackage[numbers]{natbib}

\definecolor{LightBlue}{rgb}{0.88,0.9,0.9}

\title{\Large $\ $\\ \bf Injecting Categorical Label and Syntactic Information into Biomedical NER
}

\author{\large Sumam Francis$^{*1}$ and  Marie-Francine Moens$^{1}$}
\address{\footnotesize $\ $\\$^1$ LIIR, Department of Computer Science, KU Leuven, Belgium\\

\bigskip
$^*$corresponding author
}

\abstract{\small \textit{ Biomedical Named Entity Recognition, Attribute Injection, Deep Learning, Text Classification} \normalsize
\\[17pt]
{\bf Abstract.} We present a simple approach to improve biomedical named entity recognition (NER) by injecting categorical labels and Part of Speech (POS) information into the model.  We use two approaches, in the first approach, we first train a sequence-level classifier to classify the sentences into categories to obtain the sentence-level tags (categorical labels). The sequence classifier  is modeled as an entailment problem by modifying the labels as a natural language template. This helps to improve the accuracy of the classifier. Further, this label information is injected into the NER model.  In this paper, we demonstrate effective ways to represent and inject these labels and POS attributes into the NER model. In the second approach, we jointly learn the categorical labels and NER labels. Here we also inject the POS tags into the model to increase the syntactic context of the model. Experiments on three benchmark datasets show that incorporating categorical label information with syntactic context is quite useful and outperforms baseline BERT-based models.}

\begin{document}
\thispagestyle{fancy}
\pagestyle{fancy}
\fancyhead{} 
\renewcommand{\headrulewidth}{0pt}
\fancyhead[L]{\small \texttt{Proceedings of the 18th Conference on Computational Intelligence\\Methods for Bioinformatics \& Biostatistics (CIBB 2023)}}
\fancyfoot{} 
\fancyfoot[C]{\thepage}

\section{\bf Introduction}
\label{sec:SCIENTIFIC-BACKGROUND}

Biomedical NER is a crucial task in healthcare that involves identifying and classifying entities in text into predefined categories. It is usually the first important step in solving various tasks that support clinical information extraction and decision-making. The biomedical domain requires the identification of complex entities that are not quite common in other domains. Most recent works in NER make use  of large pretrained biomedical language models like BioBERT~\cite{biobert,francis2019transfer},  ClinicalBERT~\cite{huang2019clinicalbert}, etc. which are finetuned and transferred for NER tasks. These methods only use features from raw clinical texts and learn the context of the usage of words and vocabulary in general. Incorporating attribute information such as categorical labels and syntactic information like POS can further improve the NER task performance by enhancing its context.

Previous works in biomedical NER have incorporated syntactic knowledge like POS and dependency relations by inserting these as additional attributes into the word and text representations ~\cite{tian2020improving,zhang2022biomedical}. Several other works have included attribute features as a bias term in attention mechanisms and leveraging key-value memory networks to model meaningful relations between the input words and attributes ~\cite{amplayo2019rethinking,mabert}. In this paper, we look into the task of NER to inject additional information like categorical tags and POS as attributes into the model. Both this information is quite easy to obtain since sentence-level categories (categorical labels) can be extracted from the dataset itself and POS tags can be generated by off-the-shelf NLP tools.  We inject both label and POS attributes at different locations in BERT architecture and demonstrate the effectiveness of one over the other.  We find that biasing the model through a combination of categorical label information and syntactic information leads to better recognition of entity slots.

\label{sec:DATA-AND-METHODS}

\begin{figure}[h]
\vspace{3mm}
 \begin{center}
 \includegraphics[width=0.6\textwidth]{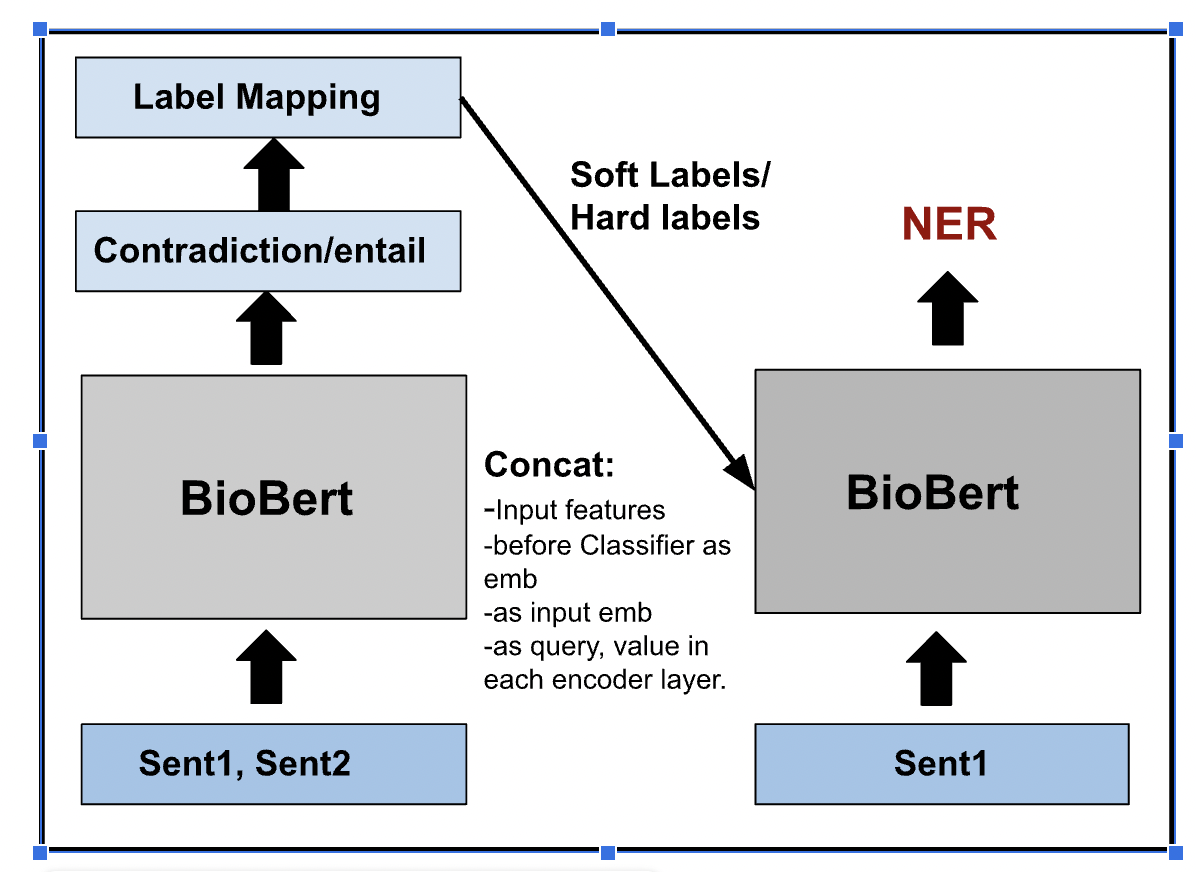}
\caption{Overview of the model: Sentence level attribute labels injected  into NER.
\label{fig:overview}}
 \end{center}
\vspace{-8mm}
\end{figure}

\section{\bf Methodology}
In this section, we describe our methodology which consists of two approaches. The first approach (pipeline approach) involves a sentence-level classifier and an attribute-injected NER. The first step is training a classifier to generate categorical labels at the sentence level including the count information. The count information indicates the number of entities present in the sentence (one or more). The second step in our training procedure is to perform NER incorporating these additional attribute features (label and POS attributes). For example: For an input sentence, \textit{A Case of Sudden Cardiac Death due to Pilsicainide - Induced Torsades de Pointes}, the categorical label is \textit{Disease} indicating the sentence level tags with \textit{Disease-more} meaning more than one entity present in the sentence and \textit{Disease-one} indicating only one entity is present. The labeling of NER labels follows BIO tagging where \textit{B} indicates the beginning, \textit{I} denotes the inside, and \textit{O} denotes outside an entity, i.e. It is not part of a biomedical entity.  We can inject attribute representation into different parts of the model. This section describes what it means to inject attributes into a certain location. The second approach (joint training) involves jointly training the sentence-level categorical tags and NER objectives to enhance the NER performance together with injecting POS attributes into different parts of the model to enhance syntactic and sentence-level context. The following sections describe each component in detail.

\subsection{\bf Sentence classifier}
In this paper, the text classification task is reformulated as a textual entailment task. The main idea is to convert the class tag into a natural language template  that describes the tag and examines if the input sentence entails the tag description \cite{wang2021entailment}. For example, a text classification (input, tag) pair: \textit{[x: A Case of Sudden Cardiac Death due to Pilsicainide - Induced Torsades de Pointes, y: Disease]} is reformulated as the following textual entailment sample: \textit{[x: A Case of Sudden Cardiac Death due to Pilsicainide - Induced Torsades de Pointes[SEP] It contains a disease mention [EOS], y: entailment]}. Reformulating the classification task as an entailment task further reduces the gap between the  Masked Language model (MLM) pre-training and fine-tuning by further pre-training the language model with an entailment task. 

\subsection{\bf Attribute injected NER}

We utilize the categorical labels obtained from the sentence classifier along with the POS as additional attributes into the Attribute injected NER. We can inject attribute representation at multiple locations of the model. This section describes what it means to inject attributes into a certain location and explores different ways and locations to represent attributes in the model. In each part of the model, we see some form of non-linear functions incorporated, using the general equation $g(f(x)) = g(W x + b)$, where $f(x)$ refers to a transformation function of $x, g$ is a non-linear activation function, $W$ and $b$ is weight matrix and bias parameters, respectively. We can represent the attributes as the bias $b$ to one of these locations by modifying them to accept sentence-level inputs ($s$) and POS tags ($p$)  as inputs, i.e. $ f(x, s, p)$.

\textbf{\bf Base Model:}
Biomedical NER is a sequence labeling task such that given an input sentence of $n$ words $X = [x_1, x_2, . .,x_n]$, the output is a sequence of named entity labels $Y = [y_1, y_2, . ., y_n]$.  The named entity labels are BIO-tagged labels labeled at the token level. For each $x_i$, the goal is to predict the corresponding label $y_i$.  The input sentence $X$ is fed into the encoder (pre-trained BioBERT) to obtain the hidden representation. Further, this is fed into a fully connected layer followed by a softmax layer. The loss function is cross-entropy.
NER label predictions are dependent on predictions for surrounding words. It has been shown that structured prediction models can improve NER performance \cite{crf}, such as conditional random fields (CRF). Here we add CRF for modeling NER label dependencies, on top of the BioBERT model.

\textbf{\bf Inject attributes into word and text representations:} 
Injecting attributes to the word embedding means that we bias the probability of a word belonging to a particular class label (obtained from sentence level classifier) independent from its neighboring context. Further injecting the Part of Speech (POS) syntactic features for each word can bias this probability further. For example, suppose an input word belongs both to a particular class label and a proper noun (NNP) POS tag, the combined attribute-injected word embeddings have a higher probability of being an entity of interest.
For the BERT-based models, we modify the  BERT model that allows us to add extra attributes as input in addition to the tokens. Here we allow the model to process additional information on the token level. This is achieved by having an extra embedding layer for the extra attributes. The full embedding vector representation of an input token is then obtained by concatenating the embeddings of the token and the additional attributes added. 

Injecting attributes to the text means that we concat the label information along with input text as a piece of additional context information.

\textbf{\bf  Inject attributes as a bias into the attention mechanism:} By injecting attributes to the attention mechanism, we bias the selection of more informative words during pooling. Adding attribute features as a bias term help to capture meaningful relations between the words and attributes. To represent the attributes through the bias parameter $b$ in the attention mechanism, the original bias $b$ is updated to $b’ = W_ss + W_pp + b$. This help to add an attribute-specific bias to the function. We inject attributes into Query ($Q$) and Key ($K$) to calculate attention scores. Instead of incorporating these attributes only into the attention mechanism as bias terms, these attributes are also injected into the input token representations.

\textbf{\bf Inject attributes into the  NER  classifier:} By injecting the additional attributes into the NER classifier, we bias the probability distribution of the NER classifier  based on the final encoded hidden representation. When we represent the categorical label and POS as a bias in the NER classifier, this produces a  biased logit vector that classifies a token as a particular entity by shifting the final probability distribution of the label class.

\subsection{\bf Joint training label and NER}
To simultaneously learn both sentence-level categories and NER tags given an input sentence $x$, BioBERT can be easily extended to a joint label classification and NER model ~\cite{jointbert}. Depending on the hidden state of the [CLS] token (special token placed at the beginning), denoted $h_1$, the sentence label is predicted as $y_i = softmax(W_ih_1 + b_i)$. For NER, we feed the final hidden states of other tokens $h_2, . . . , h_T$ into a softmax layer to classify over the NER labels $y_s$.  The learning objective is to maximize the conditional probability $p(y_i, y_s|x)$. The model is finetuned end-to-end by minimizing the cross-entropy loss. Further to enhance the syntactic context of the model we inject POS attributes to locations as described in section 2.2.  Here we add CRF for modeling NER label dependencies, on top of the joint BioBERT model.

\section{\bf Experiments and Results}
\label{sec:RESULTS}
\subsection{\bf Datasets}
We evaluate our methods on 3 biomedical benchmark datasets: the BC2GM dataset for gene/protein NER, the BC5CDR-disease dataset, and the NCBI-disease dataset for disease NER.

BC2GM ~\cite{bc2gm}:  is a dataset used for the BioCreative II gene mention tagging task. It contains 20,000 sentences from biomedical abstracts published and is annotated with 24,583 mentions of genes and proteins.

BC5CDR-disease ~\cite{bc5cdr}: is a dataset used for the BioCreative V Chemical Disease Relation (CDR) Task. It consists of  1500 titles and abstracts that are extracted from PubMed. 

NCBI-disease ~\cite{dougan2014ncbi}: contains 793 PubMed abstracts that are annotated with disease mentions and their corresponding concepts. There are 6,892 disease mentions from 790 unique disease concepts in this dataset.
We follow the study of ~\cite{biobert} to pre-process all datasets.

\subsection{\bf Implementation Details} The baseline method is a BERT-based model without attributes injected. Then we present a set of BERT-based methods incorporating categorical label information and adding syntactic features like POS using different strategies.  

In all our experiments, we use off-the-shelf NLP toolkits publicly available to generate syntactic information namely POS labels. We use Stanford CoreNLP Toolkits (SCT) to obtain the POS tagging  of a given input sentence.
For the encoder, we use the variant of BERT for the biomedical domain, i.e., BioBERT in our method. Specifically, we use  the base version of BioBERT and follow the hyper-parameters used by Lee et al. (i.e., for BioBERT-Base, there are 12 self-attention heads with 768-dimensional hidden vectors). All parameters of  the encoder are fine-tuned during training. The embeddings of attributes are randomly initialized ensuring their dimension matches the  hidden vectors dimensions of the BioBERT model. For all models, the AdamW optimizer was used with a base learning rate of 5e-5 with a warmup linear schedule. An early stopping  strategy with the patience of 5 epochs was also applied to avoid overfitting. For each method, we train 3 different models with multiple random seeds to initialize the parameters of the model and use the average of their micro F1 scores for evaluation. We train each model for 30 epochs. During each run, we evaluate the model on the validation set after each 500 steps to select the  best-performing model.
We use the same hyper-parameter settings for sentence classifier and joint BioBERT model training.

\subsection{\bf Comparative Results and Discussion} 

Table 1 shows the comparative results of different methods for biomedical NER. For models without attributes injected, the BioBERT-base model outperforms the BERT-base model \cite{devlin2018bert} mainly because it is adapted to the biomedical domain. The BioBERT models showed improvement on all datasets compared to the conventional BERT-based models due to the domain adaptation achieved through pre-training. Even in the case of BioBERT, a lack of additional features with syntactic and label knowledge  resulted in performance lower compared to the attribute-injected model variants.  Incorporating the sentence-level label together with count and POS attributes improved the performance of the model. We see the influence of using categorical labels as the results clearly indicate a better performance when combining POS together with sentence-level labels as opposed to using POS attributes only for NER.  We see that injecting attributes always results in increased performance compared to the baseline model. The best location to inject attributes from our experiments is in the embedding and concatenating with the encoder representation (input to the classifier).  

\begin{table}[httb!] \small
\centering
    \caption{Attribute injected NER results on biomedical datasets using BioBERT model.
    \label{tab:RESULTS}}
    \begin{tabularx}{0.88\textwidth}{ l  l  l l}
        \toprule
          \textbf{Method} & \textbf{BC2GM (F1)} & \textbf{BC5CDR (F1)}  & \textbf{NCBI-Disease (F1)} \\
        \midrule
         \rowcolor{LightBlue} Baseline NER (BioBERT) & $83.92\pm{0.27}$ & $86.13\pm{0.17}$ & $88.32\pm{0.38}$ \\
         \rowcolor{LightBlue} Syntax\_BioNER \cite{tian2020improving} & $84.47\pm{0.15}$ & \_ & $88.74\pm{0.26}$ \\      
         Inject in text (+ POS) & $84.26\pm{0.13}$ & $86.38\pm{0.37}$ & $88.39\pm{0.20}$ \\
         \rowcolor{LightBlue} Inject in text (+ Label) & $84.51\pm{0.17}$ & $86.46\pm{0.18}$ & $88.43\pm{0.37}$ \\
         Inject in classifier(+ POS)    &  $84.49\pm{0.15}$ &  $86.43\pm{0.33}$ & $88.81 \pm{0.18}$\\
        \rowcolor{LightBlue} Inject in classifier(+ Label)    &  \textbf{85.06}$\pm{0.19}$ &  $86.49\pm{0.25}$ & \textbf{89.01}$\pm{12}$ \\
        Inject in embedding (+ POS)&  $84.30\pm{0.16}$ & \textbf{86.89}$\pm{0.14}$ & $88.52\pm{0.23}$\\
        \rowcolor{LightBlue} Inject in embedding (+ Label)&  \textbf{84.59}$\pm{0.16}$ &  \textbf{86.98}$\pm{0.28}$ & $88.77\pm{0.10}$\\
        Inject in attention (+ POS)&  $84.35 \pm{0.15}$ &  $86.69\pm{0.22}$ & $88.63\pm{0.37}$\\
        \rowcolor{LightBlue} Inject in attention (+ Label)&  $84.59\pm{0.14}$&  $86.78\pm{0.29}$ & \textbf{88.89}$\pm{0.19}$\\
        \bottomrule
    \end{tabularx}
\end{table}

Table 2 shows the results of the joint learning of categorical labels and NER tags with a bioBERT model. Here POS tags are injected into the embedding layer to provide syntactical information to the model to learn both the sentence-level representation and token-level representation. From Table 2 we see that jointly learning categorical labels together with NER results in an increase in the performance of the model compared to the baseline NER model. Further, injecting POS labels in the joint framework can further enhance the results. This can be attributed to the fact that giving categorical labels enhance the knowledge of the model to have certain entities at the sentence level which in combination with POS tags also gives the syntactic context to improve the NER recognition especially when dealing with confusing entities.

\begin{table}[httb!] \small
\centering
    \caption{Joint learning of label information and NER results on biomedical datasets.
    \label{tab:RESULTS2}}
    \begin{tabularx}{0.85\textwidth}{ l  l  l l}
        \toprule
          \textbf{Method} & \textbf{BC2GM (F1)} & \textbf{BC5CDR (F1)}  & \textbf{NCBI-Disease (F1)} \\
        \midrule
        \rowcolor{LightBlue} Baseline NER (BioBERT) & 83.92 & 86.13 & 88.32 \\
         Joint NER &  84.93 & 86.88 & 88.84 \\
        \rowcolor{LightBlue} 
        Inject in embedding(+POS)&  \textbf{85.02} &  \textbf{86.91} & \textbf{88.99}\\
        \bottomrule
    \end{tabularx}

\end{table}


\section{\bf Conclusion}
\label{sec:CONCLUSIONS}

In this paper, we discussed two approaches to enhance BioNER with categorical label information and  syntactic information (i.e., POS labels). Our method discriminatively leverages the label information together with syntactic information to avoid the error caused by the direct use of syntactic results. The experimental results on three biomedical benchmark datasets demonstrate that syntactic information together with categorical labels can be a good resource to improve BioNER. Both the pipeline approach and joint training approach outperform the strong baseline using BioBERT. This work can be further enhanced with external knowledge from biomedical knowledge graphs.

\section*{\bf Conflict of interests}
\label{sec:CONFLICT-OF-INTERESTS}
The authors  declare there are no potential conflicts of interest.

\footnotesize
\bibliographystyle{IEEEtranN}
\bibliography{bibliography_CIBB_file.bib} 
\normalsize

\end{document}